%% file: main.tex
\documentclass[sigconf,authorversion,nonacm]{acmart}

\AtBeginDocument{%
  \providecommand\BibTeX{{%
    \normalfont B\kern-0.5em{\scshape i\kern-0.25em b}\kern-0.8em\TeX}}}

\copyrightyear{2022}
\acmYear{2022}
\setcopyright{rightsretained}
\acmConference[ICMI '22]{INTERNATIONAL CONFERENCE ON MULTIMODAL INTERACTION}{November 7--11, 2022}{Bengaluru, India}
\acmBooktitle{INTERNATIONAL CONFERENCE ON MULTIMODAL INTERACTION (ICMI '22), November 7--11, 2022, Bengaluru, India}
\acmDOI{10.1145/3536221.3556596} 
\acmISBN{978-1-4503-9390-4/22/11}

\usepackage{multirow}
\usepackage{booktabs}

\begin{document}

\title{Generalized Product-of-Experts for Learning Multimodal Representations in Noisy Environments}

\author{Abhinav Joshi}
\email{ajoshi@cse.iitk.ac.in}
\affiliation{
  \institution{IIT Kanpur}
  \city{Kanpur}
  \country{India}
}

\author{Naman Gupta}
\authornote{Both authors contributed equally to this research.}
\email{namang@cse.iitk.ac.in}
\affiliation{
  \institution{IIT Kanpur}
  \streetaddress{IIT Kanpur}
  \city{Kanpur}
  \country{India}
  \postcode{208016}
}

\author{Jinang Shah}
\authornotemark[1]
\email{jinang.iitk@gmail.com}
\affiliation{
  \institution{IIT Kanpur}
  \streetaddress{IIT Kanpur}
  \city{Kanpur}
  \country{India}
  \postcode{208016}
}

\author{Binod Bhattarai}
\authornote{Corresponding author}
\email{b.bhattarai@ucl.ac.uk}
\affiliation{
  \institution{University College London}
  \city{London}
  \country{U.K.}
}

\author{Ashutosh Modi}
\authornote{Both authors were senior supervisors}
\email{ashutoshm@cse.iitk.ac.in}
\affiliation{
  \institution{IIT Kanpur}
  \streetaddress{IIT Kanpur}
  \city{Kanpur}
  \country{India}
}

\author{Danail Stoyanov}
\authornotemark[3]
\email{danail.stoyanov@ucl.ac.uk}
\affiliation{
  \institution{University College London}
  \streetaddress{Gower St}
  \city{London}
  \country{U.K.}
}


\input{abstract}

\keywords{Multimodal Representations; Multimodal Fusion; Cross-modal Processing; Deep Learning Architectures; Machine Learning}

\maketitle
\input{Introduction}

\input{RelatedWorks}

\input{Methodology}

\input{Experiments}

\input{Results-Analysis}

\input{Conclusion}

\section{Acknowledgments}
We would like to thank reviewers for their insightful comments. 

\noindent Ashutosh Modi is supported in part by SERB India (Science and Engineering Board) (SRG/2021/000768). 

\noindent Binod Bhattarai and Danail Stoyanov are funded by in whole, or in part, by the Wellcome/EPSRC Centre for Interventional and Surgical Sciences (WEISS) 
(203145/Z/16/Z), Engineering and Physical Sciences Research Council (EPSRC)
(EP/P012841/1), the Royal Academy of Engineering Chair in Emerging Technologies scheme, and EndoMapper project by Horizon 2020 FET (GA863146).

\bibliographystyle{ACM-Reference-Format}
\bibliography{references}
\clearpage

\end{document}

%% file: abstract.tex
\begin{abstract}

A real-world application or setting involves interaction between different modalities (e.g., video, speech, text). In order to process the multimodal information automatically and use it for an end application, Multimodal Representation Learning (MRL) has emerged as an active area of research in recent times. MRL involves learning reliable and robust representations of information from heterogeneous sources and fusing them. However, in practice, the data acquired from different sources are typically noisy. In some extreme cases, a noise of large magnitude can completely alter the semantics of the data leading to inconsistencies in the parallel multimodal data. In this paper, we propose a novel method for multimodal representation learning in a noisy environment via the generalized product of experts technique. In the proposed method, we train a separate network for each modality to assess the credibility of information coming from that modality, and subsequently, the contribution from each modality is dynamically varied while estimating the joint distribution. We evaluate our method on two challenging benchmarks from two diverse domains: multimodal 3D hand-pose estimation and multimodal surgical video segmentation. We attain state-of-the-art performance on both benchmarks. Our extensive quantitative and qualitative evaluations show the advantages of our method compared to previous approaches.

\end{abstract}

%% file: Introduction.tex
\section{Introduction}

Humans interact with the real world by conveying and perceiving information using multiple modalities. For example, when two people talk to each other, besides the primary modality of verbal communication via context (text), they also use additional modalities like the tone of speech (audio) and facial/hand gestures (video). The information from these varied modalities may either overlap or complement each other. Moreover, the signals captured in the real world are often noisy. Humans in real-world interaction may not be able to capture signals from all the modalities efficiently, often leading to noisy signals. However, humans tend to fuse the noisy information efficiently and learn about one modality from another modality. For example, the emotions of a person speaking in an unknown language can be predicted using their voice tone and physical gestures without any context information. Based on this intuition, researchers have focused on developing methods for exploiting shared information between the different modalities for self-supervision and combining the complementary information to improve a machine learning model's generalization capability. The area of \textit{Multimodal Representation Learning} (MRL) involves fusing the information coming from varied sources to learn representations that are robust and generalizable in different settings (for example, the case of a missing modality). However, the existing methods in MRL often tend to assume the training dataset to be noise-free. In this work, we propose a model that efficiently handles the noise present in different modalities in the dataset and compare it with the existing modality fusing mechanisms.

In recent years, generative models like Variational Auto-Encoders (VAE)s have attracted colossal interest in modality fusing mechanisms \cite{ lee2021private, shi2020relating, majumder2019variational, geenjaar2021fusing}
. The ability of VAEs to create information bottlenecks for the available modalities makes it easier to fuse information coming from different sources. The learned unimodal posteriors are combined using a fusing mechanism, for example, some of the widely popular fusing mechanisms include Product of Experts (PoE)~\cite{wu2018multimodal}, and Mixture of Experts (MoE)~\cite{shi2019variational}. PoE multiplies unimodal posteriors, while MoE sums them up. Both of these approaches have their own merits and demerits, and we refer the reader to ~\cite{sutter2020generalized} for details. In particular, compared to MoE, PoE can aggregate any subset of modalities, which provides an efficient way of dealing with the missing modalities. This property has attracted its usage in various tasks such as multimodal 3D hand-pose estimation~\cite{Cai_2018_ECCV}, and multimodal-GAN to generate naturally realistic images~\cite{huang2021multimodal}. In this paper, we, too, use a variant of a PoE method for learning multimodal representations in a noisy setting.

Another noteworthy component in mixing information from multiple modalities is dynamically deciding the importance of a modality sample.
Humans tend to figure out the noisy signals from a specific modality easily and fuse the information by giving less weightage to the modalities with the noisy signals. For example, a person parking a vehicle with a faulty rear-view camera (noisy image) will rely more on the rear-view mirrors for making parking decisions. Or a person communicating via video conferencing where the audio signals are noisy will shift their attention towards lip reading from the visuals to understand the speaker efficiently. Dynamically deciding the credibility of a sample from a modality makes the fusion of multimodal information more efficient and robust in such noisy cases. Considering the importance of dynamic weightage in modality fusing mechanisms, we propose to estimate the weightage contribution for each of the modalities in our architecture.

\begin{figure}[t]
\centering
  \includegraphics[scale=0.36]{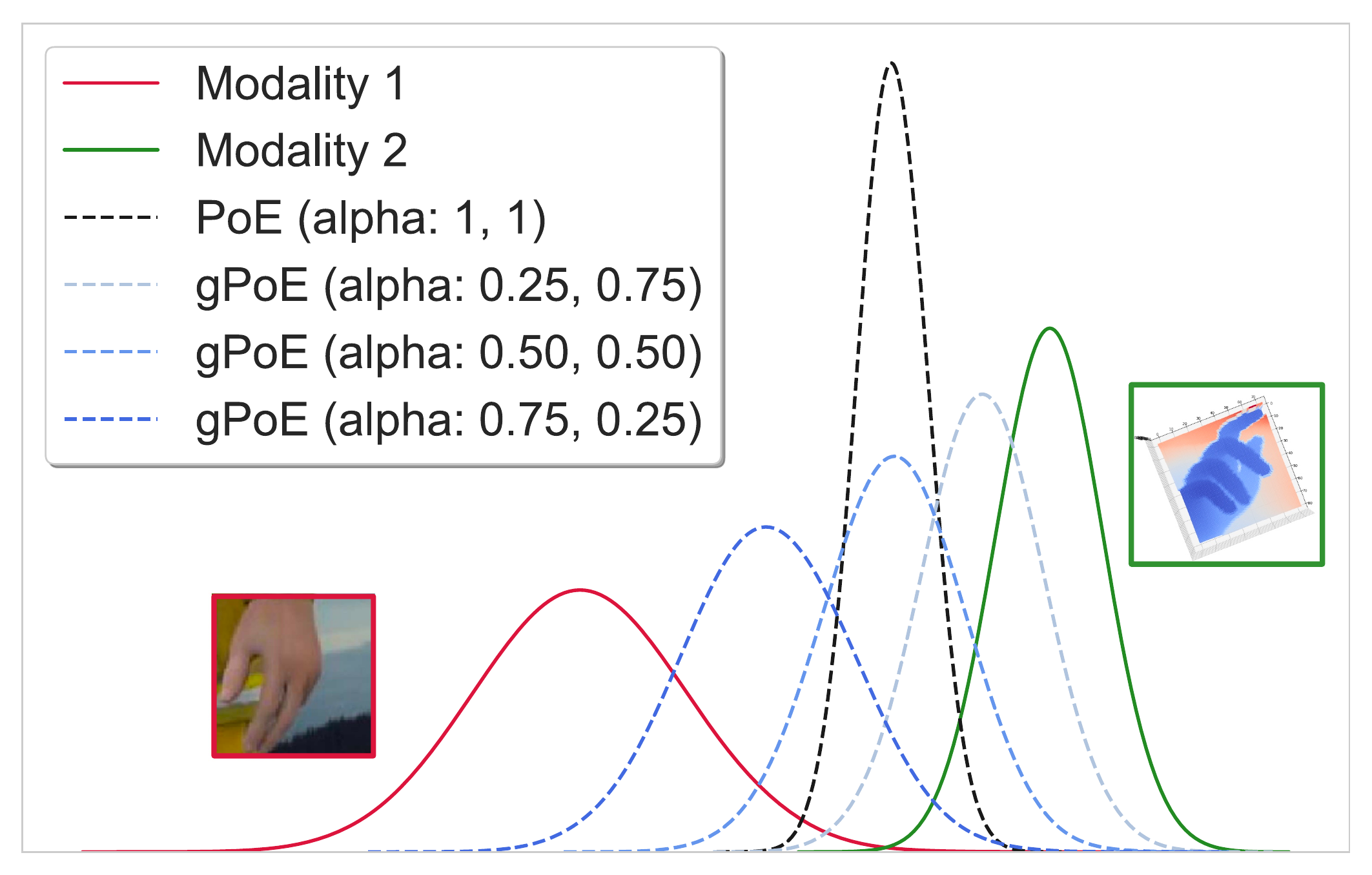}
\caption{The figure shows a comparison between the joint distribution obtained by Product-of-Experts (PoE) and Generalized Product-of-Experts (gPoE) while fusing two modalities. The alpha introduced in gPoE helps scale the unimodal posteriors ({red} and {green} curves), allowing the freedom to appoint different weightage to different modalities dynamically. The blue curves show the multiple distributions obtained with varying weightage (alpha) values. The darker blue shade (alpha: 0.75, 0.25) gives more weightage to Modality 1 ({red}) whereas the lighter blue shade (alpha: 0.25, 0.75) gives more weightage to Modality 2 ({green}).}
  \label{fig:dist}
\end{figure}

PoE multiplies unimodal posteriors assuming a  uniform contribution from every modality. It results in a peaky joint distribution if all marginal posteriors have high density and vice-versa, as shown in Figure \ref{fig:dist}, where, the green and the red curves show the distribution of two different modalities, and a PoE combination gives a dashed black curve. This is an optimal combination if the information from each modality is equally credible. However, in a real-world scenario, this may not hold. Various types of equipment/sensors, such as depth sensors, RGB cameras, event cameras, LIDAR, etc., are used to capture different modalities. Equipment encounters different levels of noise asynchronously. The varying noise in each source corrupts it differently, and sometimes it can destroy the semantics of that modality and thus affecting the combination of modalities.
Consequently, simply combining the posterior of each modality can be detrimental and eventually deteriorate the performance. Thus, there is a need to assess the credibility of information in every constituent modality, reweigh the parameters based on the credibility, and fuse them. For example, the effects of reweighing of unimodal posteriors on their joint posterior are shown in Figure \ref{fig:dist} via the dashed blue curves. The demo\footnote{\href{https://www.desmos.com/calculator/l4y75hedez}{https://www.desmos.com/calculator/l4y75hedez}} shows how introducing the alpha parameter to PoE helps reweigh the contribution of two modalities, allowing the freedom to appoint different weightage to different modalities dynamically.

Inspired by recent work on re-scaling Gaussian processes~\cite{corr_CaoF14}, we propose training different networks for each modality to check the credibility of the data coming from each source. In particular, we propose a cross-modal VAE based architecture (\S\ref{sec:methodology}{}) for learning latent representation for each modality and then using a generalized product of experts for taking a weighted combination of the latent representations. We evaluate the proposed method on two tasks: multimodal hand pose estimation and multimodal semantic segmentation in surgical videos. Since both datasets are synthetic, to emulate a real-world setting, we add different noise levels to different modalities and perform an extensive set of experiments on the corrupted input. Our results show improvements over the existing  PoE-based methods and attain state-of-the-art performance on the tasks.

%% file: RelatedWorks.tex
\section{Related Works}


\noindent\textbf{Multimodal Generative Learning:}
For the case of two modalities, several variants of variational autoencoders (VAEs) \cite{kingma2014autoencoding,NIPS2014_d523773c} have been proposed to learn generative models for a uni-directional conditional inference. However, we are more interested in approaches that learn joint latent space for better modeling of data distribution, which can be used for conditional inference interchangeably across all modalities. Joint multimodal variational autoencoder (JMVAE) proposed in \citet{suzuki2016joint} attempts to learn a joint distribution explicitly but by training separate inference networks for each possible subset of present modalities, which becomes intractable as more modalities get involved. \citet{wu2018multimodal} introduces multimodal variational autoencoder (MVAE), which leverages Product-of-Experts (PoE), whereas Multimodal Mixture-of-Experts VAE (MMVAE), as proposed by \citet{shi2019variational}, uses a Mixture of Experts (MoE) to learn joint distribution efficiently. These approaches only measure their robustness in terms of their ability to handle missing modalities during inference, which is necessary but, in turn, do not attempt to address their limitations against noisy multimodal input.  \\

\noindent\textbf{Dealing with Noise:} 
To address the issue of noisy inputs in multimodal settings, numerous approaches have been proposed \cite{9413164,DBLP:journals/corr/abs-2002-09708,DBLP:journals/mia/SudarshanUECA21} for specific tasks, but they largely remain restricted to discriminative models. Thus, leaving a gap in research to address limitations of current approaches focused on learning a joint distribution in the presence of noisy multimodal inputs. Since PoE has been widely used and documented for multimodal generative networks across various tasks, we use PoE to investigate further how noisy inputs limit the performance for specific tasks of 3D Hand Pose Estimation and Semantic Segmentation.  \\

\noindent\textbf{3D Hand Pose Estimation:}
Generative methods attempt to learn the distribution of the hand model from the given observations and remain highly susceptible to initialization. Although using depth or 3D data (especially 3D point clouds \cite{Li_2019_CVPR,8578976}) provides the most accurate results \cite{DBLP:journals/corr/OberwegerWL15,DBLP:journals/corr/abs-1711-07399}, their availability usually remains uncertain during training or inference. Thus, recent works \citet{Yang_2019_ICCV,Cai_2018_ECCV,spurr2018crossmodal} leverage depth information for training while restricting testing exclusively with RGB images. While \citet{Cai_2018_ECCV} utilizes rendered depth maps from poses to regularize RGB image-based training, \citet{spurr2018crossmodal} proposes a VAE-based method to learn a shared latent space, which suffers from its alternating training strategy for different modalities and a relatively slower convergence. \citet{Yang_2019_ICCV} proposes a multimodal VAE-based method, where, authors rather attempt at aligning latent spaces of diverse modalities, including 3D poses, point clouds, and heat maps using PoE. Rather than forcing learning of a joint latent space, aligning latent spaces from different modalities allows a much faster convergence and also a better handling with non-corresponding data from other available modalities. \\

\noindent\textbf{Semantic Segmentation:}
As semantic segmentation plays a vital role in medical diagnosis and treatment, the particular domain has been well studied. In recent years, there have been numerous deep learning-based approaches suggested to utilize the multimodal data for medical image segmentation as detailed in \citet{ZHOU2019100004}. Yet, semantic segmentation using multimodal generative learning and its analysis with noisy inputs remains an understudied area for the medical domain. We, thus, consider a recently published Surgical Video-Sim2Real dataset \cite{Rivoir_2021_ICCV} for semantic segmentation with multiple modalities, which includes RGB images and depth maps. More details about the dataset are mentioned in (\S\ref{sec:Datasets}{}).

%% file: Methodology.tex
\section{Methodology} \label{sec:methodology}

For our setting, we consider a data acquisition pipeline with multiple sensors for capturing information tuples (e.g., audio, video, text ) emanating from multiple sources. Since information from every modality can be noisy, it can lead to corruption of the sample. Consider the acquisition of a sample instance $i$ having $M$ different modalities: $\{\mathbf{m}^{(i)}_1, \mathbf{m}^{(i)}_2, \ldots, \mathbf{m}^{(i)}_M\}$. Consider the probability of noise in respective sensors to be $\{p_{\mathbf{m}_1}, p_{\mathbf{m}_2}, \ldots, p_{\mathbf{m}_M} \}$, with increasing number of sensors, the probability of multiple noisy modalities  occurring simultaneously decreases exponentially $\prod_{k=1}^M p_{\mathbf{m}_k} \ll 1$. In contrast, the number of corrupted data samples increases with the number of sensors. The primary assumption of clean, noise-free data samples from all the modalities limits the current deep learning architectures to use corrupted data samples when even one modality is noisy. Recently, a wide variety of approaches (using MVAE \cite{wu2018multimodal}) have tried to efficiently handle the missing modalities by aligning the learned latent spaces. However, the major limitation of such an approach is the underlying assumption that the available modalities are always noise-free.

Moreover, PoE giving equal weightage to all modalities results in a distribution dominated by highly confident modality-specific experts. For noisy examples, the experts produced by the encoders might not represent the required task-specific underlying data distribution. For such noisy cases, it is better to rely on noise-free experts. To control the credibility of each of the modalities, we introduce the use of \textit{Generalized Product-of-Experts} \cite{corr_CaoF14}, where the information present in different modalities controls the contribution of an expert dynamically. In the next section, we describe the architecture details. We start with the formulation of standard-VAE and extend it to crossmodal domain. Further, we introduce the use of PoE for mixing information from multiple modalities. In the end, we formulate the generalized PoE that dynamically captures every modality's contribution to help learn multimodal representations in a noisy environment effectively.

\subsection{Architecture}

\noindent \textbf{Standard VAEs} are encoder-decoder based generative architectures that maximise the evidence lower bound (ELBO) of the data log-likelihood. The amortized variational inference scheme with reparameterization trick \cite{kingma2014autoencoding,VI-with-normalizing-flows} helps formulating the approximate posterior $q(\phi(\mathbf{z}|\mathbf{x}))$ and the likelihood $p_{\theta}(\mathbf{x} | \mathbf{z})$ distributions using deep neural networks with parameters $\phi$ and $\theta$ respectively.
\begin{equation}
\begin{aligned}
  \mathcal{L}(\theta, \phi) &= \mathbb{E}_{\mathbf{z}\sim q_\phi(\mathbf{z}|\mathbf{x})}[\log p_\theta(\mathbf{x}|\mathbf{z})] 
  - \mathcal{D}_{KL}[q_\phi(\mathbf{z}|\mathbf{x}) \| p(\mathbf{z})]  
\end{aligned}
\end{equation}
where $\mathcal{D}_{KL}$ represents the Kullback-Leibler divergence and $p(\mathbf{z})$ represents a prior distribution which can vary from a standard Gaussian \cite{kingma2014autoencoding} to more expressive priors like normalizing flows \cite{kingma2017improving,grathwohl2018ffjord,berg2018sylvester}.

\noindent \textbf{Crossmodal-VAE:} A standard VAE can be extended for modelling crossmodal inference, i.e. translating information from one modality to another. Consider two modalities, $\mathbf{m}_{input} = \{\mathbf{x}^{(1)}, \mathbf{x}^{(2)}, \ldots, \mathbf{x}^{(n)}\}$ and $\mathbf{m}_{target} = \{\mathbf{y}^{(1)}, \mathbf{y}^{(2)}, \ldots, \mathbf{y}^{(n)}\}$, the ELBO in crossmodal VAE is minimized to generate the target modality ($\mathbf{y}^{(i)}$) from its corresponding pair in input modality ($\mathbf{x}^{(i)}$).
\begin{equation}
\begin{aligned}
\mathcal{L}(\theta, \phi) &= \mathbb{E}_{\mathbf{z}\sim q_\phi(\mathbf{z}|\mathbf{x})}[\log p_\theta(\mathbf{y}|\mathbf{z})] - \mathcal{D}_{KL}[q_\phi(\mathbf{z}|\mathbf{x}) \| p(\mathbf{z})]
\end{aligned}
\end{equation}
Further, for our case we assume the presence of other auxiliary modalities $\mathbf{M}_{aux} = \{\mathbf{m}_1, \mathbf{m}_2, \ldots, \mathbf{m}_N\}$ during training and extend the minimization objective for crossmodal VAE. Thus, we additionally have $\mathbf{M}_{aux} = \{\mathbf{m}_1, \mathbf{m}_2, \ldots, \mathbf{m}_N\}$, and the total number of available modalities becomes $N+2$ including the primary input and output modalities.
\begin{equation}
    \mathbf{M} = \{\mathbf{m}_1, \mathbf{m}_2, \ldots, \mathbf{m}_N, \mathbf{m}_{input},  \mathbf{m}_{target}\}
\end{equation}

\begin{figure*}[t]
\centering
  \includegraphics[width=0.95\textwidth]{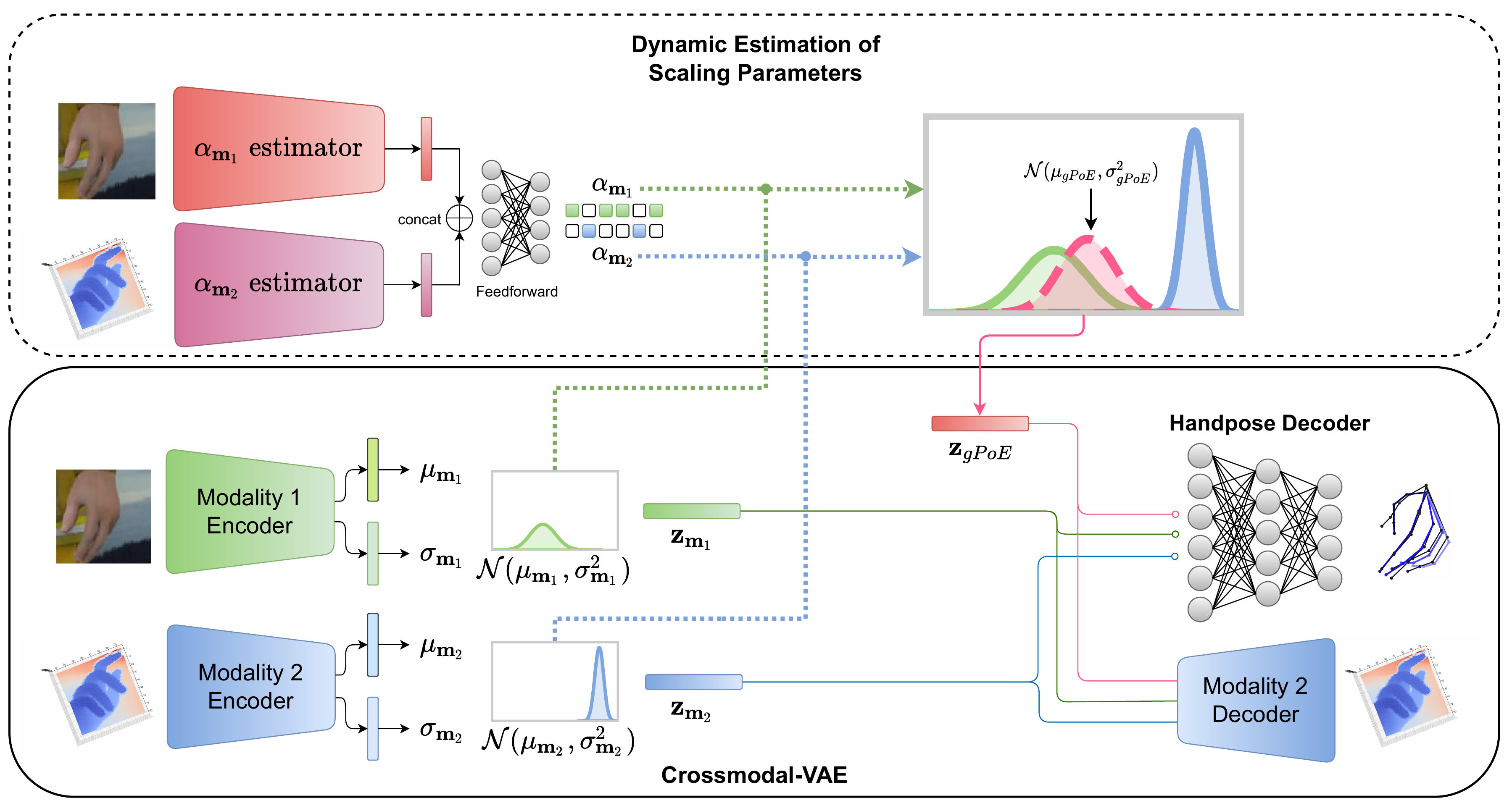}
  \caption{Crossmodal-VAE architecture with gPoE used for 3D hand pose estimation. The (top) estimated scaling parameters $\{\alpha_1,\alpha_2\}$ are used with their corresponding learned modal distributions to identify the aligned distribution using gPoE as shown in Eq.\eqref{eq:8} and Eq.\eqref{eq:9}. The proposed framework although shown for two modalities here can be easily extended to any $N$ number of modalities present. }
  \label{fig:architecture}
\end{figure*}

\noindent \textbf{Generalized Product-of-Experts:} For mixing the features from multiple modalities, existing approaches \cite{wu2018multimodal, DaunhawerSMV20DMVAE} use the product of experts (PoE) for computing the joint latent representation, which is proportional to the individual modality distributions.

\begin{align}
    \label{eq:PoE_p}
    P(\mathbf{z})&=\frac{1}{Z} \prod_{i} p_{i}(\mathbf{z})
\end{align}
\begin{align}
    \label{eq:PoE_mu}
    \mu_{PoE}(\mathbf{z})&=\left(\sum_{i} \mu_{i}(\mathbf{z}) \mathrm{T}_{i}(\mathbf{z})\right)\left(\sum_{i}
    \mathrm{~T}_{i}(\mathbf{z})\right)^{-1} 
\end{align}
\begin{align}
    \label{eq:PoE_sigma}
    \Sigma_{PoE}(\mathbf{z})&=\left(\sum_{i} \mathrm{~T}_{i}(\mathbf{z})\right)^{-1}
\end{align}
where $\mathrm{T}_{i}(z)=\Sigma_{i}^{-1}(z)$ is the precision of the $i^{th}$ Gaussian expert in $\mathbf{z}$, and $Z$ is the normalization constant. PoE generates distribution dominated by highly confident experts compared to the less confident ones. The presence of noise in the dataset can cause an expert to produce erroneously low predicted variance along a latent dimension which further develops a strong bias in the joint predictions. To overcome this issue, we propose using a generalized formulation of the product of experts (gPoE) \cite{corr_CaoF14} which introduces a weighing mechanism for scaling down such overconfident experts as formulated below.
\begin{align}
    P(\mathbf{z}) &= \frac{1}{Z} \prod_{i} p_{i}^{\alpha_{i}(\mathbf{z})}(\mathbf{z})\\
    \mu_{gPoE}(\mathbf{z}) &= \left(\sum_{i} \mu_{i}(\mathbf{z}) \alpha_{i}(\mathbf{z}) \mathrm{T}_{i}(\mathbf{z})\right)\left(\sum_{i} \alpha_{i}(\mathbf{z})  \mathrm{T}_{i}(\mathbf{z})\right)^{-1} \label{eq:8}\\
    \Sigma_{gPoE}(\mathbf{z}) &= \left(\sum_{i} \alpha_{i}(\mathbf{z}) \mathrm{T}_{i}(\mathbf{z})\right)^{-1}
    \label{eq:9}
\end{align}
For handling the noisy information present in the dataset, we formulate the modality-specific scaling factors $\alpha_{i}$ as a direct function of input modalities ($\mathcal{F}(\mathbf{m}_i)$).
We estimate the scaling parameters $\alpha$ for each dimension in the latent space using independent modality encoders. An overview of our architecture is shown in Figure \ref{fig:architecture}, where features from different modalities are concatenated and passed through standard feed-forward layers. For a normalized scaling of predicted gaussian distributions, we use a softmax function to distribute the importance across available modalities such that for each latent dimension $\sum_{i=1}^{N+2} \alpha_i = 1$. Further, using gPoE, we define the joint representation of available modalities as,
\begin{equation}
\mathbf{z}_{joint} \sim  \mathcal{N}(\mu_{gPoE}, \Sigma_{gPoE})
\end{equation}

\begin{equation}
\begin{aligned}
\mathcal{L} (\Theta, \Phi) = \mathcal{L}&\bigg(\{ \theta^{(\mathbf{y})}, \theta^{(\mathbf{m}_1)}, \ldots, \theta^{(\mathbf{m}_N)}\},  
 \{ \phi^{(\mathbf{x})}, \phi^{(\mathbf{m}_1)}, \ldots, \phi^{(\mathbf{m}_N)} \} \bigg) 
\end{aligned}
\end{equation}

For generating the target modality $\mathbf{y}$, the reconstruction loss is derived as follows:
\begin{equation}
\begin{aligned} 
\label{eq:recontarget}
\displaystyle \mathcal{L}_{target} &= \mathbb{E}_{\mathbf{z}_{joint}}[\log p_\theta(\mathbf{y} \mid \mathbf{z}_{joint})] +
\mathbb{E}_{\mathbf{z}_{\mathbf{x}}}[\log p_\theta(\mathbf{y} \mid \mathbf{z}_{\mathbf{x}})] 
\\ &+
\sum_{i=1}^{N} \mathbb{E}_{\mathbf{z}_{\mathbf{m}_i}}[\log p_\theta(\mathbf{y} \mid \mathbf{z}_{\mathbf{m}_i})]
\end{aligned}
\end{equation}

Similarly, for generating other auxiliary modalities, the reconstruction loss is,

\begin{equation}
\begin{aligned}
\label{eq:reconaux}
\mathcal{L}_{aux} = \sum_{k=1}^{N} &\bigg ([\mathbb{E}_{\mathbf{z}_{joint}}[\log p_\theta(\mathbf{m}_k \mid \mathbf{z}_{joint})] \\ &+
\mathbb{E}_{\mathbf{z}_{\mathbf{x}}}[\log p_\theta(\mathbf{m}_k \mid \mathbf{z}_{\mathbf{x}})] \\ &+ 
\sum_{i=1}^{N} \mathbb{E}_{\mathbf{z}_{\mathbf{m}_i}}[\log p_\theta(\mathbf{m}_k \mid \mathbf{z}_{\mathbf{m}_i})])\bigg)
\end{aligned}
\end{equation} 
Furthermore, the Kullback-Leibler divergence for all the latent spaces is,
\begin{equation}
\begin{aligned}
    \label{eq:kld}
    \mathcal{L}_{KL} &= \mathcal{D}_{KL}[q_{\phi_\mathbf{x}}(\mathbf{z}_{\mathbf{x}} \mid \mathbf{x}) \| p(\mathbf{z})] 
    + \sum_{i=1}^N \mathcal{D}_{KL}[q_{\phi_{\mathbf{m}_i}}(\mathbf{z}_{\mathbf{m}_i}|\mathbf{m}_i) \| p(\mathbf{z})]
\end{aligned}
\end{equation} \\
Adding the losses in \ref{eq:recontarget}, \ref{eq:reconaux} \ref{eq:kld}, 
we obtain the minimising objective of our formulation
\begin{equation}
\label{eq:learning_objective}
\mathcal{L} (\Theta, \Phi) = \mathcal{L}_{target} + \mathcal{L}_{aux} + \beta \mathcal{L}_{KL}
\end{equation}
where $\beta$ is a hyperparameter \cite{higgins2016betaVAE} used to balance the loss between reconstruction and Kullback-Leibler divergence.
Note that the objective of our work is to leverage the information present in corresponding modality pairs during training that helps make unimodal inference robust towards noisy data samples.

%% file: Experiments.tex
\section{Experiments}

\subsection{Datasets}
\label{sec:Datasets}
To evaluate our method in the noisy environment, we choose two synthesized datasets from different domains, which provide multiple modalities that help formulate a unimodal prediction task. The two publicly available datasets include the Rendered Hand Pose Dataset (RHD)~\cite{zb2017hand} for predicting 3D hand pose from RGB images and Surgical Video-Sim2Real Dataset~\cite{Rivoir_2021_ICCV} for predicting segmentation masks from RGB images. \\

\noindent \textbf{RHD} is a synthesized dataset of rendered hand images with 320×320 resolution. The dataset was created from 20 subjects, performing 39 actions that were rendered using Blender \cite{blender}. 
It consists of 41258 training and 2728 testing samples. The dataset provides RGB images, along with the corresponding depth maps, segmentation masks, and 3D keypoints for hand pose prediction. \\

\noindent \textbf{Surgical Video-Sim2Real} dataset consists of 21000 randomly sampled views for 7 simulated, surgical 3D scenes, which were rendered using the liver meshes obtained from the 3D-IRCADb dataset \cite{soler_hosteller_agnus_charnoz_fasquel_moreau_osswald_bouhadjar_marescaux_2010} composed of 3D CT scans of liver. The surgical dataset provides realistic translation views and the corresponding depth maps, camera poses, and segmentation masks. The provided segmentation mask consists of 5 classes: liver, fat/stomach, abdominal wall, gallbladder, and ligament. 

\subsection{Preprocessing}
For hand pose prediction, we follow a preprocessing scheme similar to \citet{Yang_2019_ICCV} and crop the hand region of the image. Further, using the available hand segmentation mask, we extract the corresponding hand region from the depth image and convert it to point clouds using the provided camera intrinsic parameters. For segmentation prediction in the Surgical Sim2Real dataset, we center-crop the RGB image for visual modality and resize it to 256x256. The available depth images are converted to point clouds using the provided camera intrinsic parameters. 


\subsection{Noise Simulation}

\begin{figure*}[t]
    \centering
    \includegraphics[width=\textwidth]{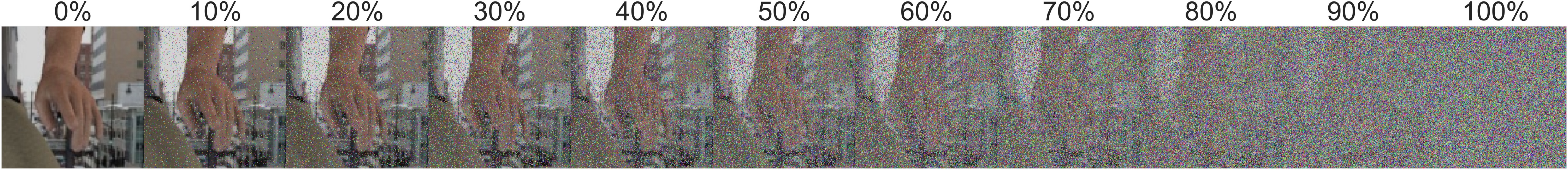}\\
    \includegraphics[trim={5.51 40.7cm 0 37.5cm},clip, width=\textwidth]{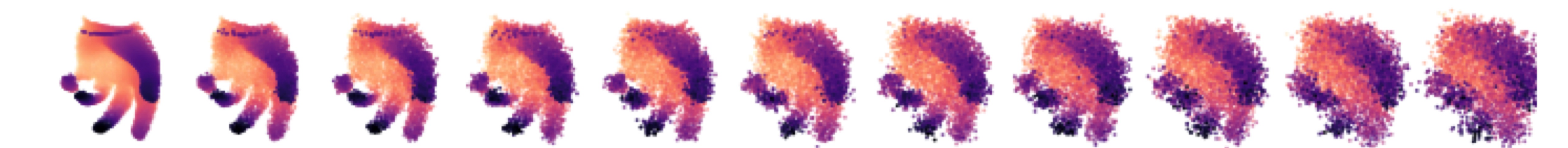}\\
    \caption{(top) Random pixel noise added to image modality. (bottom) Gaussian noise (increasing left to right) added to a point cloud.}
    \label{fig:noise}
\end{figure*}

Modeling imaging sensor noise is a fundamental problem in image processing. 
The practical application pipeline of camera sensors is highly complex and usually has different modules that cause various types of noise inductions in an acquired image. Many existing approaches have proposed statistical noise models to simulate real-world noise in the acquired images. Recently, a wide variety of deep-learning methods have been introduced to simulate the real-world noise \cite{noise-Abdelhamed2019NoiseFN, noise-chang2020learning, noise-Chen2018ImageBD, noise-ECCV2020_984, noise-Kim2019GRDNGroupedRD, noise-Marras2020ReconstructingTN}. In contrast, various methods use simple diffusion models to introduce Gaussian noise in images which are further used to make the deep learning architectures more robust towards a noisy environment. In general, noise in a visual modality can be interpreted as information loss in terms of pixels. To simulate noise for our synthetic environment, we choose a simple scheme of corrupting the pixels by replacing the pixel values with random values. Figure \ref{fig:noise} shows an example of an image with various percentages of pixel corruption added as noise.
In real-world applications, sensors like LiDAR, RADAR, etc., have shown poor performance in adverse weather conditions resulting in deviated computation of point clouds. For generalizing our methods to multiple modalities, we consider the auxiliary modality to be noisy and simulate noise using standard Gaussian noise applied to the observed values. Figure \ref{fig:noise} shows an example of Gaussian noise added to a depth point cloud sample.

\subsection{Evaluation Metrics}
For evaluation of the predicted hand poses in the RHD dataset, we use the standard metrics \cite{Yang_2019_ICCV}: mean end-point-error (EPE) and area under the curve (AUC) on the percentage of correct key points (PCK) curve (higher is better). EPE is computed using the euclidean distance between the predicted and ground-truth 3D keypoints of  hand joints (lower is better). PCK represents the percentage of predicted key points that fall within certain error thresholds of EPE. To evaluate the predicted segmentation masks in the Surgical Sim2Real dataset, we use standard IoU and F1 metrics. IoU or Jaccard similarity coefficient computes the size of the intersection divided by the size of the union of two label sets, and F1 score is interpreted as a harmonic mean of the precision and recall, where an F1 score reaches its best value at 1 and worst score at 0. 

\subsection{Implementation Details}
For a fair comparison and showing the effect of gPoE over PoE over the noisy environment, we follow the encoder and decoder architectures inspired from \citet{Yang_2019_ICCV}. We consider RGB as our primary modality and use a pre-trained ResNet-18 \cite{He_2016_CVPR} with fully connected layers to generate $\mu_{RGB}$ and $\sigma_{RGB}$. For encoding point clouds, we use the ResPEL network \cite{Li_2019_CVPR}, for decoding point clouds, we use the Folding-Net decoder \cite{yang2018foldingnet}. We use a set of linear layers in the RHD dataset to decode hand keypoints and standard DC-GAN architecture \cite{radford2016unsupervised} to predict segmentation maps in the surgical dataset, keeping the rest of the architecture the same in both the datasets.

%% file: Results-Analysis.tex
\section{Results-Analysis}
\begin{figure*}
  \centering
  \includegraphics[width=0.245\textwidth,trim={1cm 0.5cm 2cm 0cm}, clip]{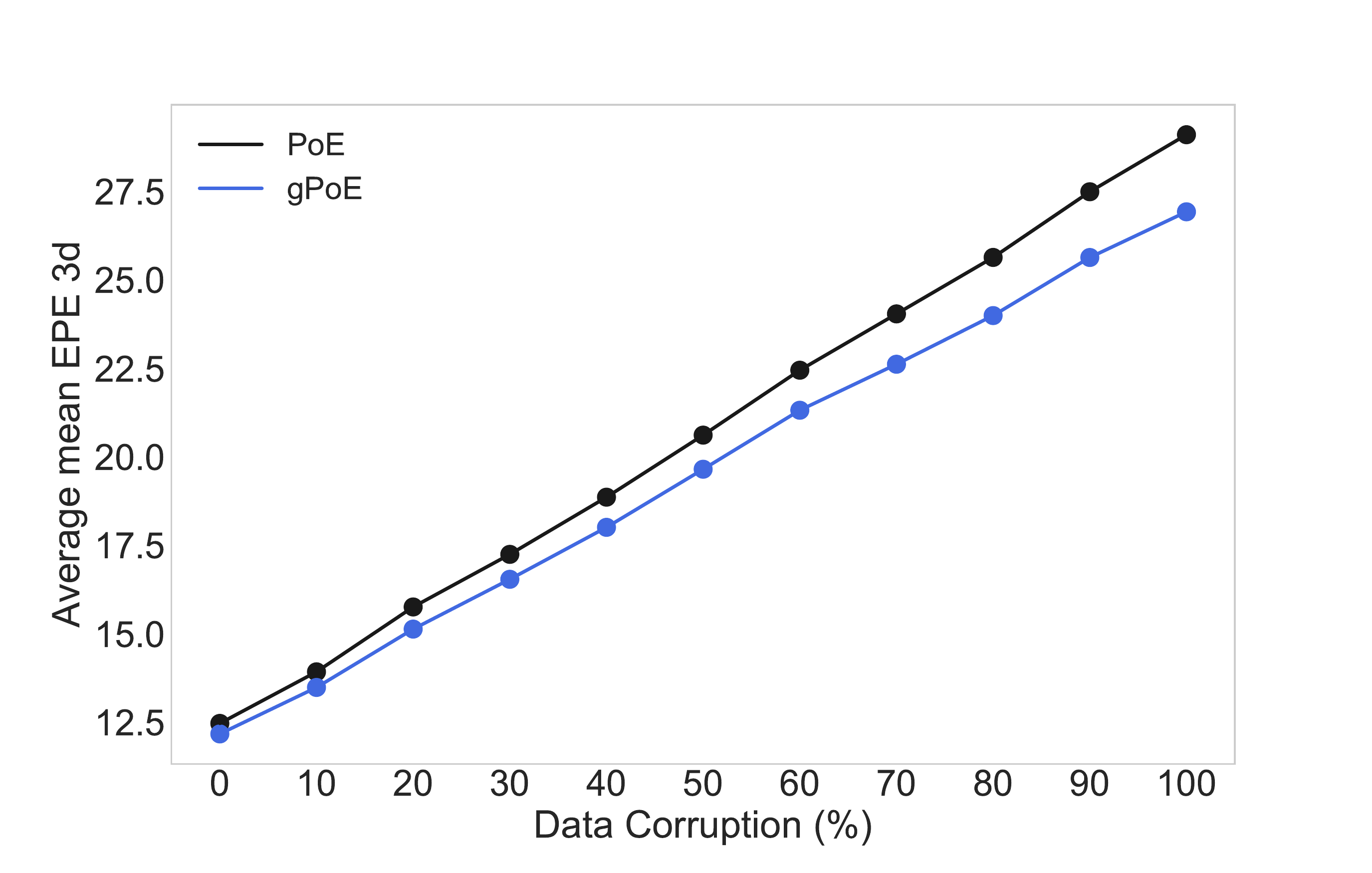}
  \includegraphics[width=0.245\textwidth,trim={1cm 0.5cm 2cm 0cm}, clip]{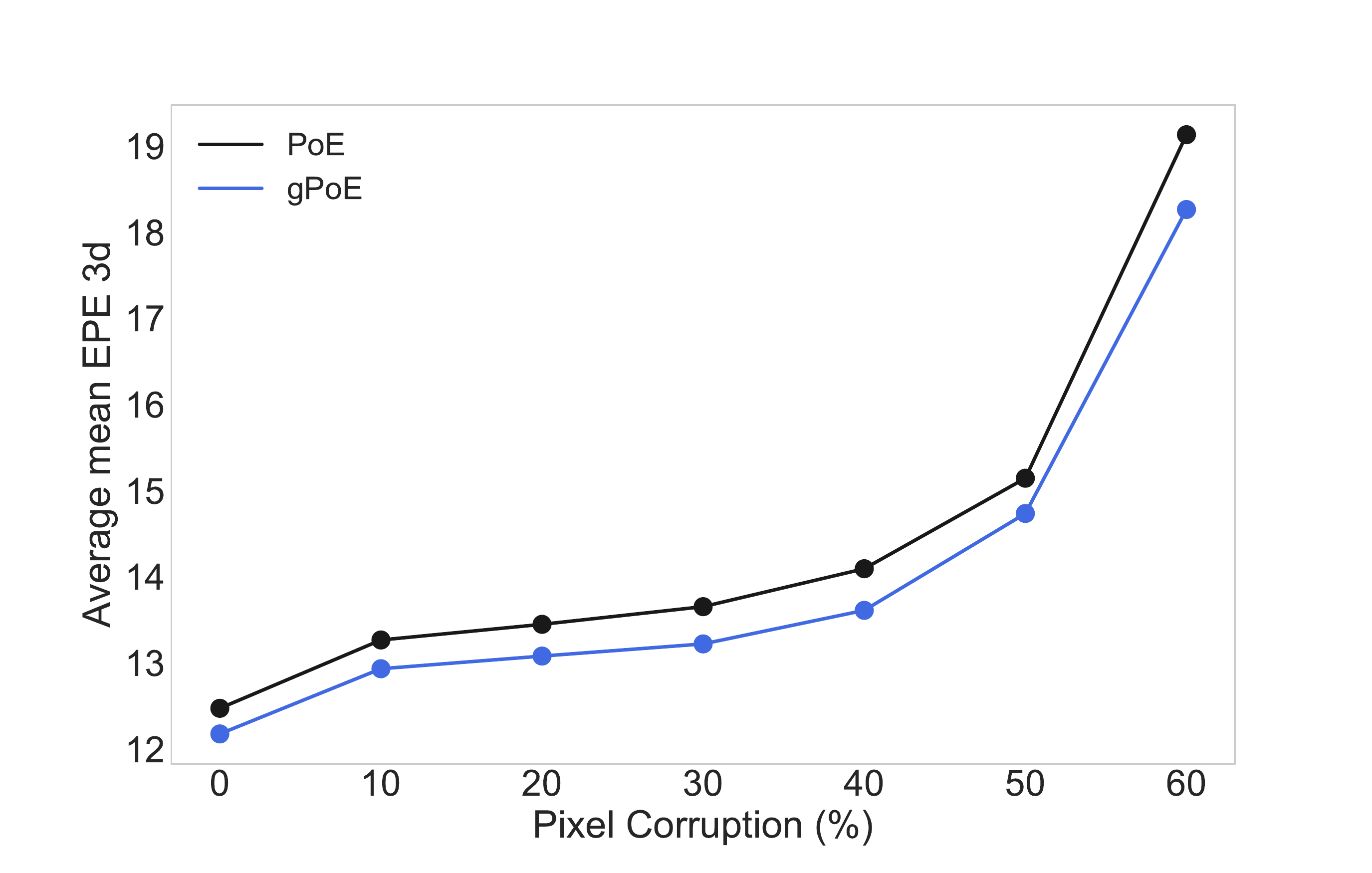}
  \includegraphics[width=0.245\textwidth, trim={1cm 0.5cm 2cm, 0cm }, clip]{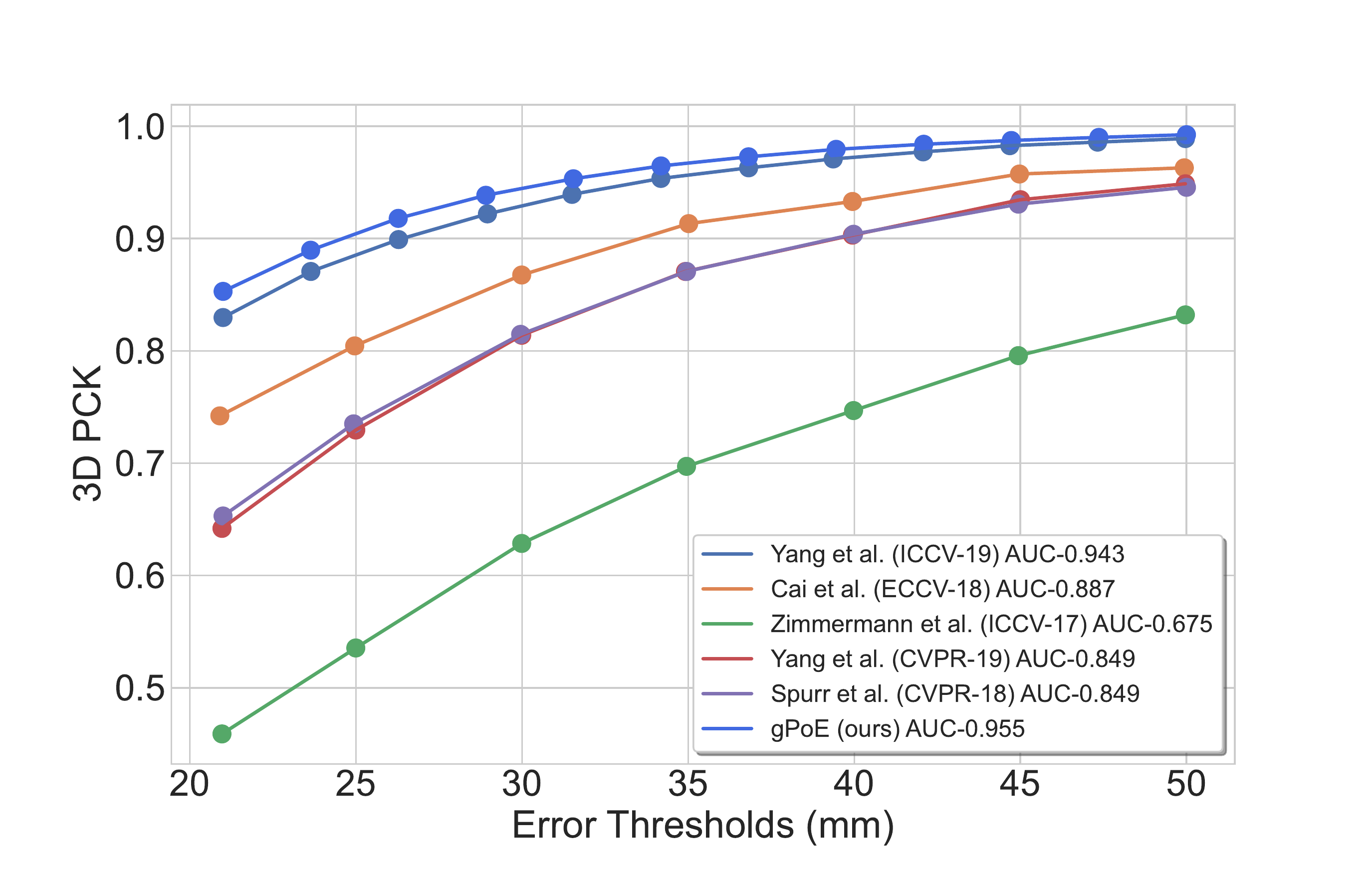}
  \includegraphics[width=0.245\textwidth, trim={1cm 0.5cm, 2cm, 0cm, clip}]{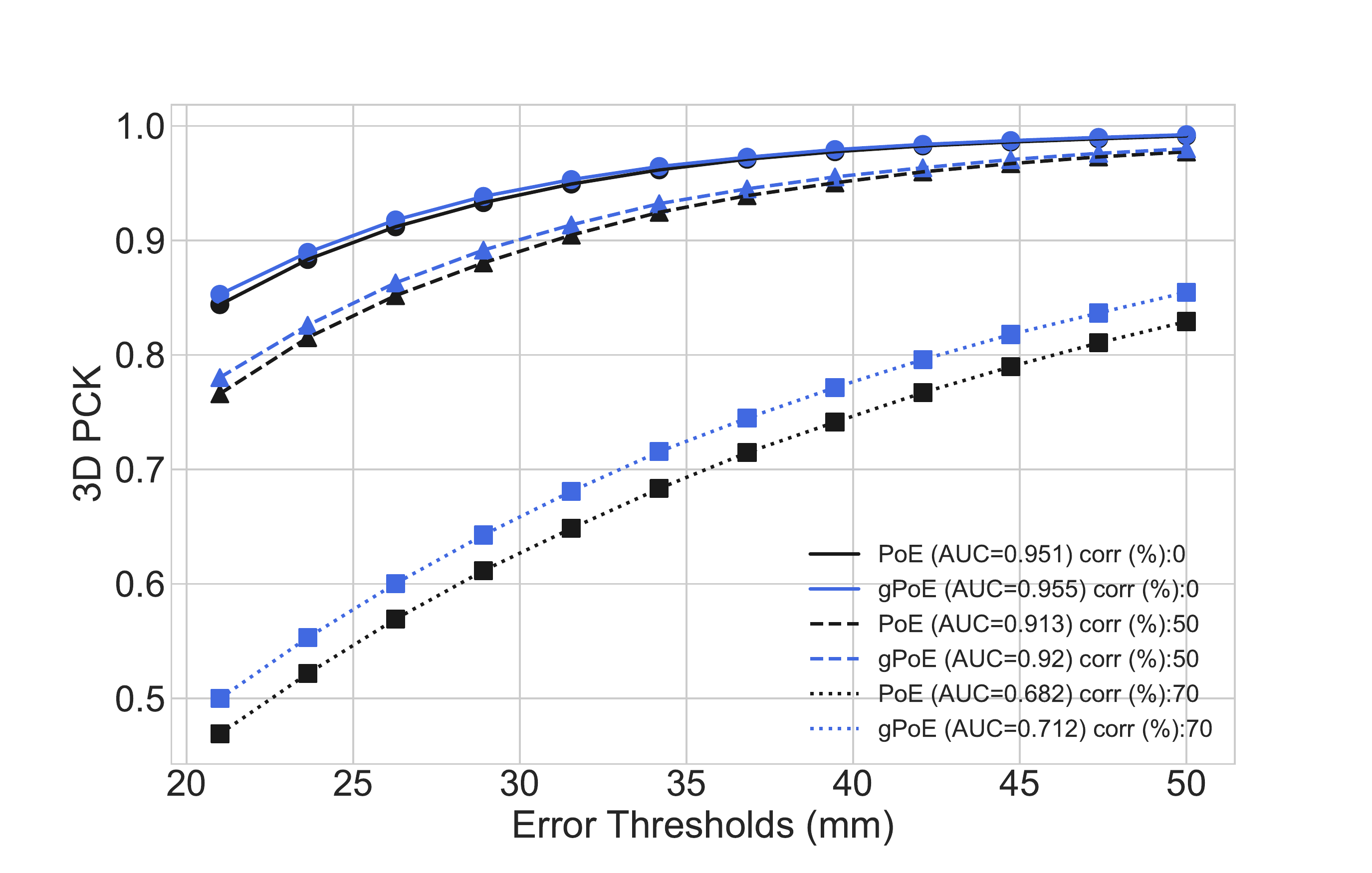}
  \caption{The first two figures from the left show the Average mean EPE 3D (mm) comparison between PoE and gPoE for varying levels of data corruption (first) and pixel corruption (second), respectively, on the RHD dataset. 
The third figure shows the 3D PCK vs. Error Thresholds (mm) plot comparison against prior works, and the fourth figure shows the performance across different noise levels on the RHD dataset  (Zoom in for a better view).}
  \label{fig:data_corruption_epe_and_baseline_comparison}
\end{figure*}

  

\begin{figure*}[!ht]
    \centering
  \includegraphics[width=0.855\textwidth]{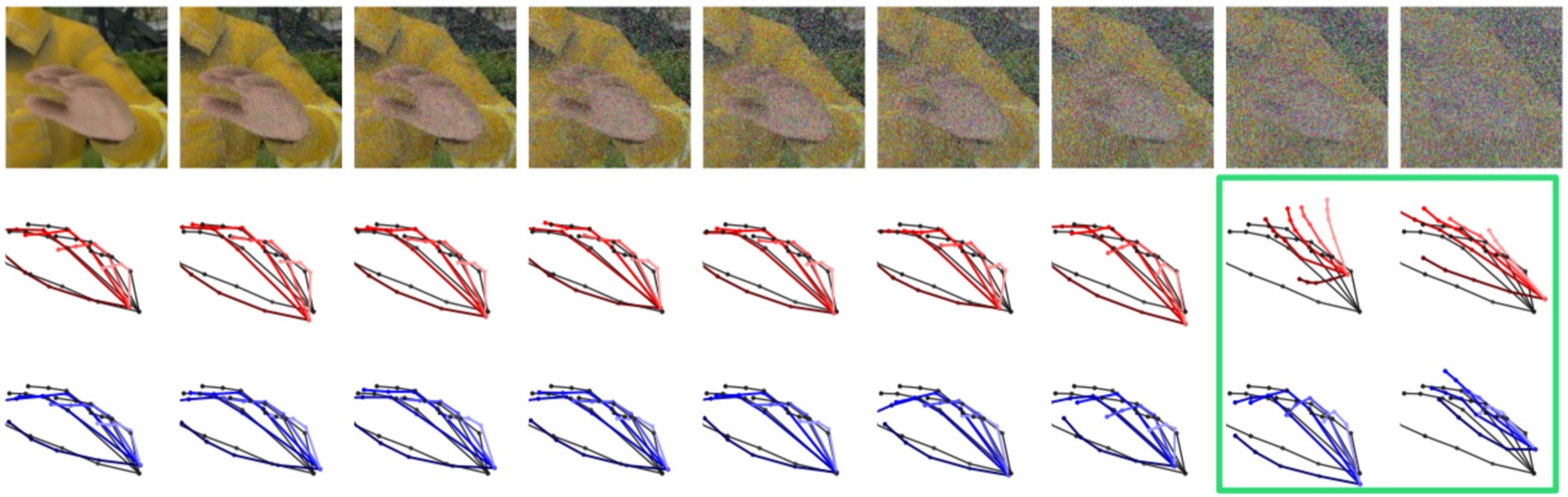}

  \includegraphics[width=0.855\textwidth]{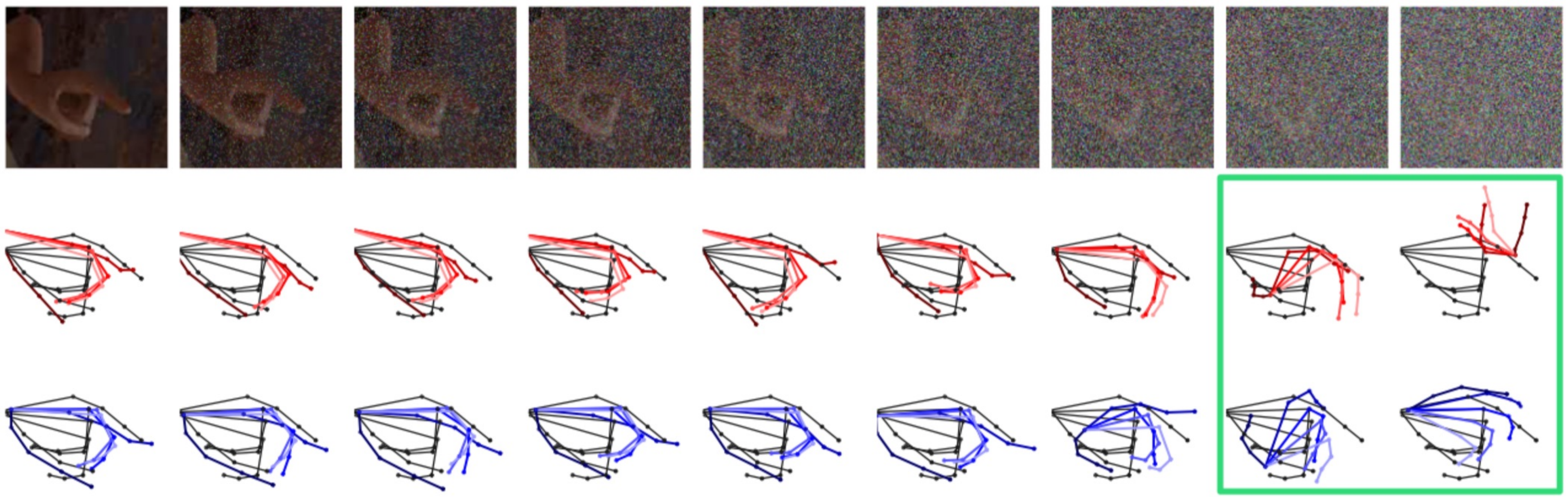}
  
  \includegraphics[width=0.855\textwidth]{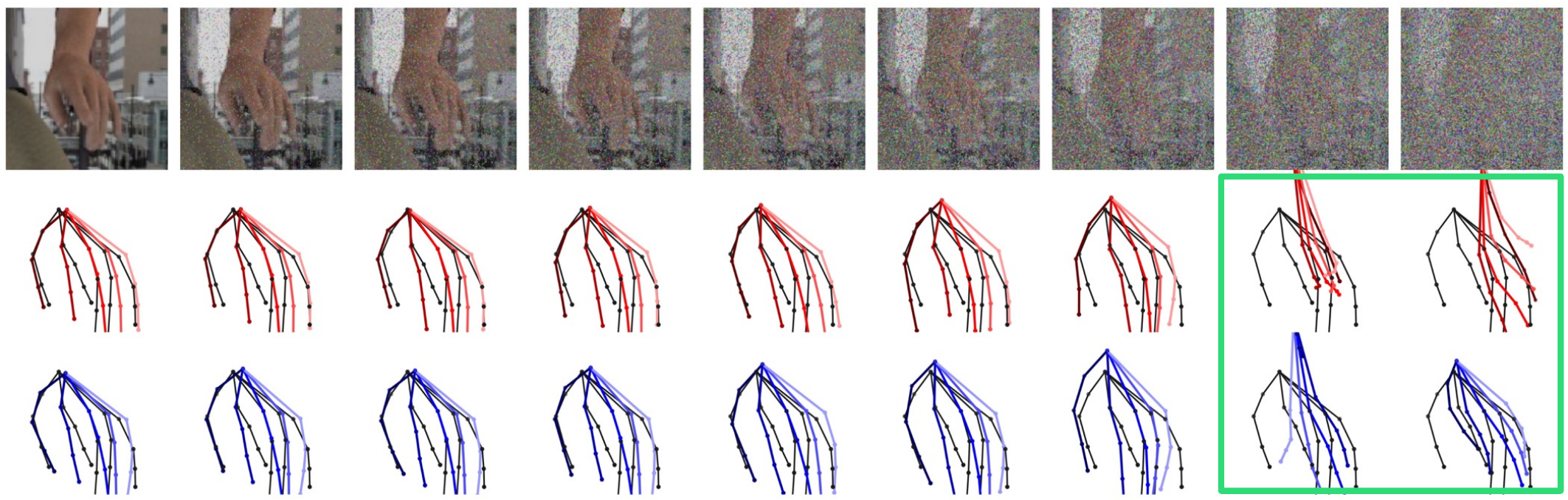}
  
  \includegraphics[width=0.855\textwidth]{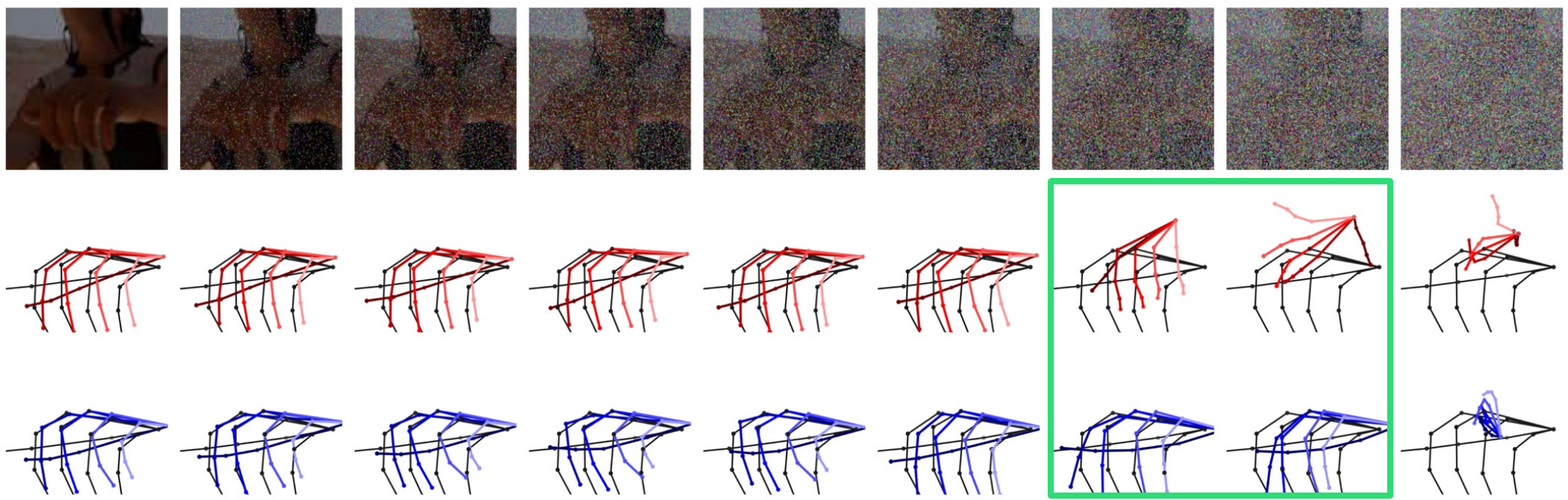}
  
   \caption{
  Hand Pose Estimation using PoE (\textcolor{red}{red}) and gPoE ( \textcolor{blue}{blue}) on the RHD dataset. 
  The first row in each sample shows the input RGB image, and the ground truth keypoints are shown in (\textcolor{black}{black}). The second and third row shows the predicted masks from PoE and gPoE, respectively.
Noise intensity increases going from left to right. Green boxes highlight significant improvement in predicted hand poses for extremely noisy inputs.
}
  \label{app:fig_handpose_inference}
\end{figure*}

\begin{figure*}[!ht]
    \centering

  \includegraphics[width=0.855\textwidth]{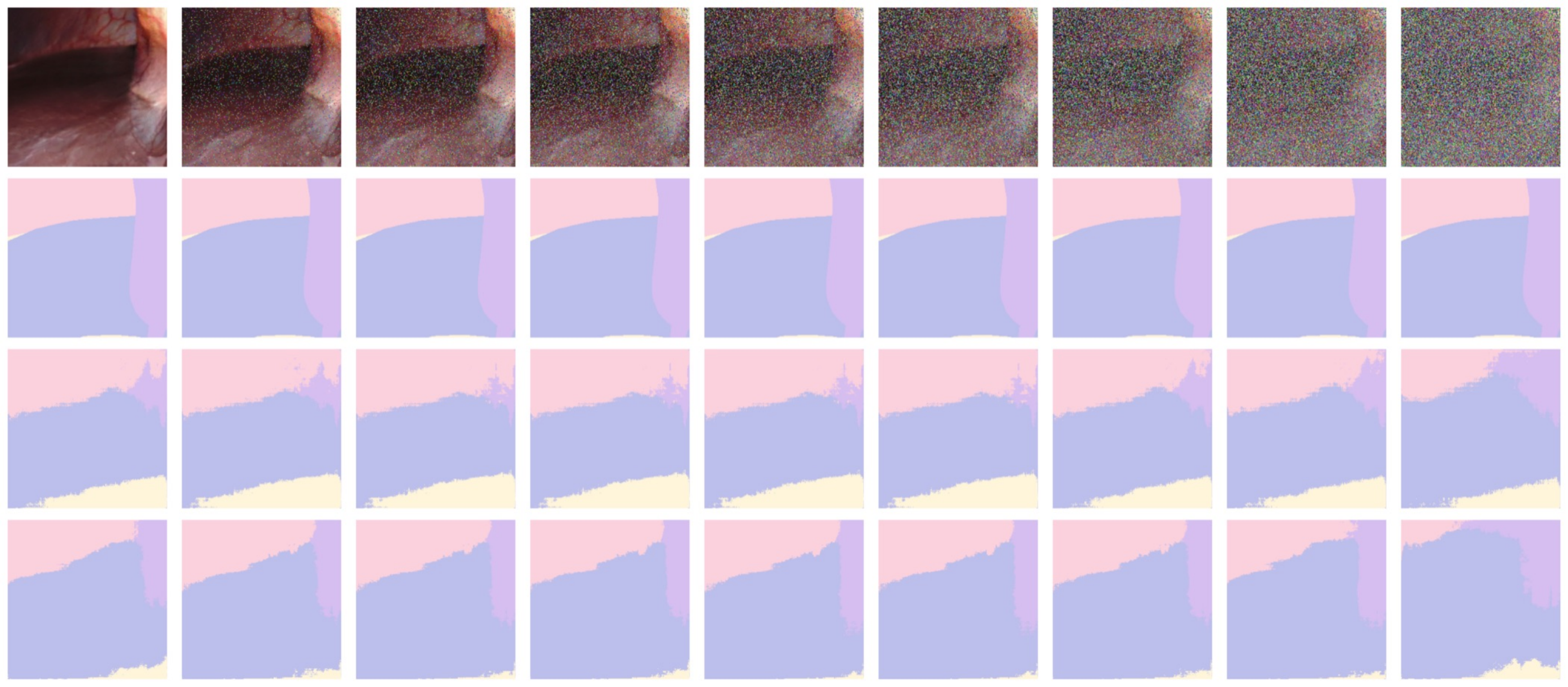}
  
  \includegraphics[width=0.855\textwidth]{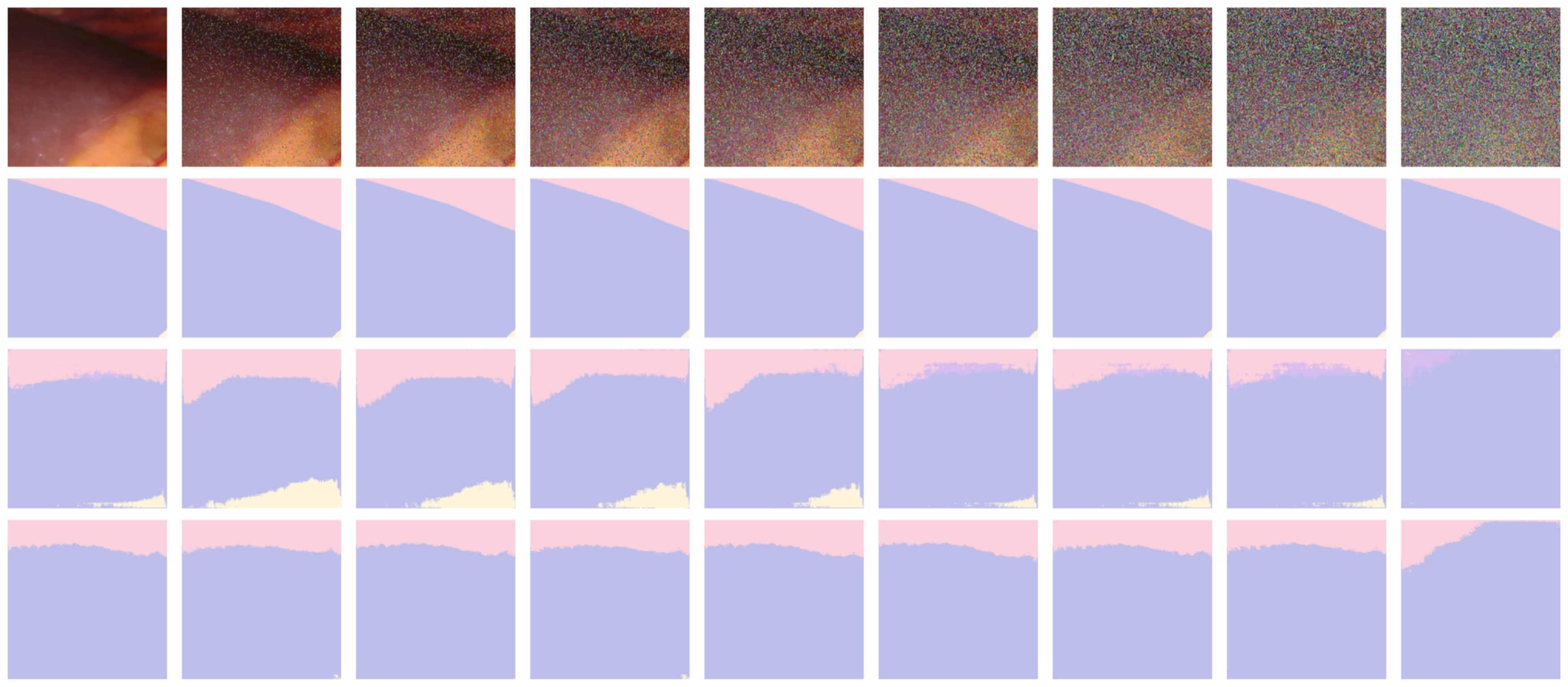}
  
  \includegraphics[width=0.855\textwidth]{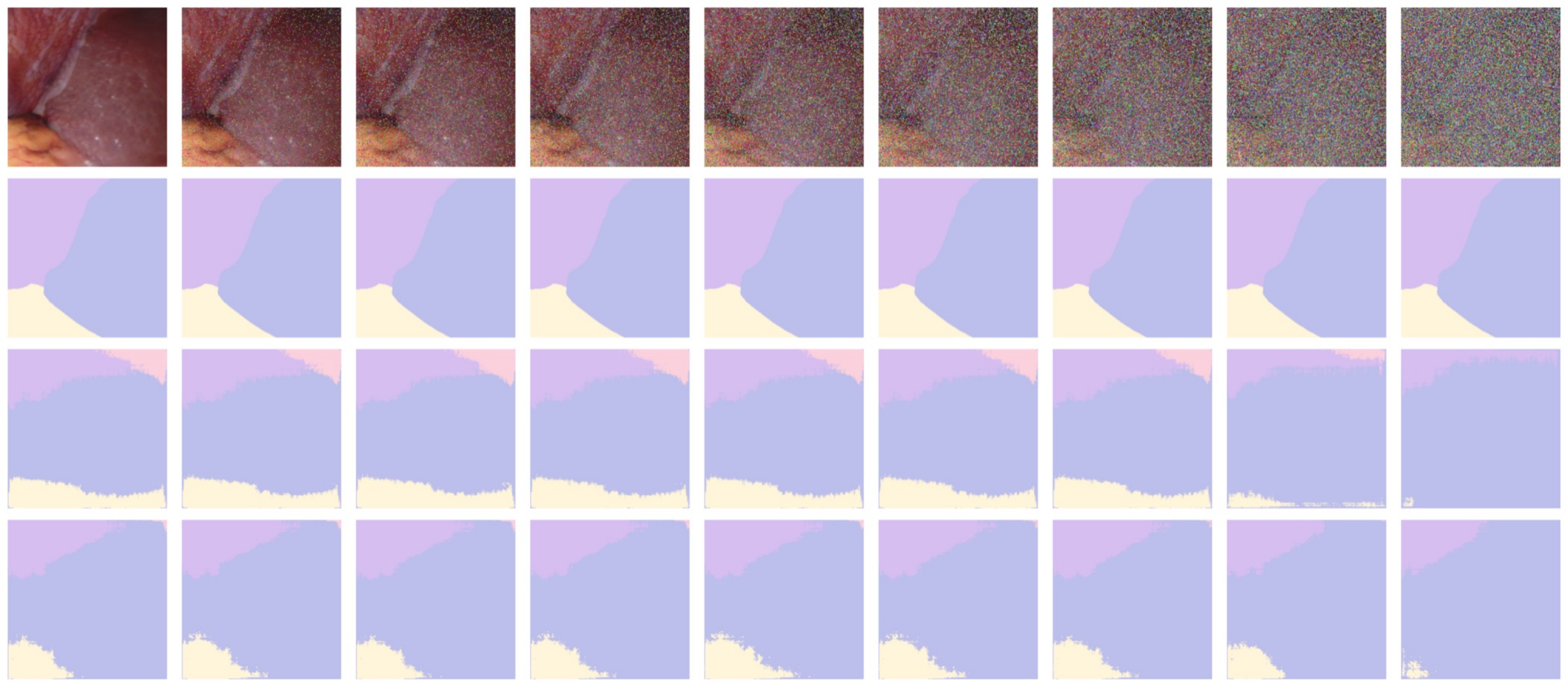}

  \caption{{Surgical Semantic Segmentation comparison on architectures trained with PoE and gPoE in a noisy setting (noise increasing from left to right). The first row in a sample represents the input image. The second row represents the ground truth segmentation map, third and fourth row shows the predicted masks from PoE and gPoE, respectively.}}
  \label{app:fig_surgical_prediction}
\end{figure*}



\subsection{Quantitative Results}

\begin{table}[t]
\renewcommand{\arraystretch}{1.1}
  \centering
  \begin{tabular}{cc@{} cc@{} cc@{} cc@{}}
    \setlength{\tabcolsep}{20pt}
    \\ \toprule
\textbf{Pixel Corruption}      & \textbf{Method} & \textbf{ IOU}          & \textbf{F1}           \\     \midrule
\multirow{2}{*}{\textbf{0\%}}  & PoE             & 0.6272          & 0.7692          \\  
                               & gPoE            & \textbf{0.7138} & \textbf{0.8323} \\ 
   \midrule
\multirow{2}{*}{\textbf{50\%}} & PoE             & 0.6221          & 0.7653          \\  
                               & gPoE            & \textbf{0.7078} & \textbf{0.8281} \\ 
    \midrule    
\multirow{2}{*}{\textbf{70\%}} & PoE             & 0.6128          & 0.7576          \\  
                               & gPoE            & \textbf{0.6866} & \textbf{0.8131} \\ 
    \bottomrule
  \end{tabular}
  \caption{Comparison of gPoE and PoE for semantic segmentation on the surgical dataset with varying degree of pixel corruption.}
  \label{tab:seg_surgical_data}
\end{table}

To validate the effectiveness of our proposed architecture in a noisy setting, we compare the performance of unimodal encoder models trained in similar noisy settings over various ranges of data corruption. The increasing deviation in performance shown in Figure \ref{fig:data_corruption_epe_and_baseline_comparison} (first from the left) represents the robustness of generalized PoE towards noisy data samples. The difference in performance increases almost linearly with the amount of data corruption depicting the performance gain when the larger proportion of the test dataset is noisy. Moreover, to judge the efficacy of the proposed architecture, we also test the unimodal encoders on a varying range of pixel corruption. For this setting, we corrupt all the test set examples with a fixed percentage of pixel corruption and observe the performance for both architectures. Figure \ref{fig:data_corruption_epe_and_baseline_comparison} (second from the left) shows the Average Mean EPE varying across the pixel corruption percentage. As observed, gPoE outperforms PoE over the entire range of pixel corruption, highlighting the generalizability of gPoE in a noisy natural environment. 

Furthermore, to quantitatively judge the effect of pixel corruption, we compare the performance of PoE and gPoE using the PCK curve for a considerable amount of pixel corruption. To show this effect, we choose 3 settings, 0\%, 50\%, and 70\% pixel corruption. Figure \ref{fig:data_corruption_epe_and_baseline_comparison} (fourth from the left) shows the observed PCK curves for these settings. The effectiveness of our method is observable when the noise in the dataset increases. The significant performance gap between the 50\% and 70\% pixel corruption highlights the adverse effects of noise in the test environment. The increasing difference in AUC values between gPoE and PoE (3\% difference in 70\% pixel corruption) shows the robustness of our system in a noisy environment. The encoder-decoder models trained with gPoE are less prone to noise for unimodal inference. In contrast, the encoder-decoder trained with PoE performs inadequately when pixel corruption increases in the test set.

We also test our approach by experimenting with the Surgical Video-Sim2Real dataset to predict pixel-level segmentation masks using RGB images and corresponding depth information. In Table \ref{tab:seg_surgical_data}, we provide a comparison between PoE and gPoE for three different levels of pixel corruption on the surgical dataset for semantic segmentation. In this setting, gPoE outperforms PoE by a significant margin showing the robustness towards noisy samples present during training. The introduction of dynamic scaling in gPoE not only helps in training the unimodal encoders simultaneously but also results in unimodal encoders, which are more robust against real-world noisy data samples.

\subsection{State-of-the-Art Comparison:}
\begin{table}[t]
\renewcommand{\arraystretch}{1.1}
\setlength\tabcolsep{11pt}
  \centering
  \begin{tabular}{@{}lc@{}rc@{}}
    \toprule
    \textbf{Method} & \textbf{mean EPE (mm)} & \textbf{AUC}  \\
    \midrule
    Zimmerman et al. \cite{zimmermann2017learning} (ICCV-17) & - & 0.675\\
    Spurr et al. \cite{spurr2018crossmodal} (CVPR-18) & 19.73 & 0.849\\
    Cai et al. \cite{Cai_2018_ECCV} (ECCV-18) & - & 0.887\\
    Yang et al. \cite{yang2019disentangling} (CVPR-19) & 19.95 & 0.849 \\
    Yang et al. \cite{Yang_2019_ICCV} (ICCV-19) & 13.14 & 0.943\\
    \textbf{gPoE (ours)} & \textbf{12.18} & \textbf{0.955}\\
    \bottomrule
  \end{tabular}
  \caption{Comparing gPoE with prior works for 3D Hand Pose Estimation on RHD dataset. Lower mean end-point-error (EPE) is better, higher AUC is better.}
  \label{tab:baseline_comparison}
\end{table}

Though the main focus of our work is tackling a realistic noisy setting, for fair analysis,  we briefly highlight the comparison with the other existing works on hand pose prediction from RGB images. To validate our results on the RHD dataset, we compare our method with the existing 3D hand pose prediction approaches. Spurr et al. \cite{spurr2018crossmodal} and Yang et al. \cite{Yang_2019_ICCV} are related to our method as they use VAE based architectures for hand pose prediction. Out of the two approaches, Yang et al. \cite{Yang_2019_ICCV} come closer to our approach and act as our baseline for highlighting the drawbacks of the product of experts in a noisy environment. Table \ref{tab:baseline_comparison} compares the related works on 3D hand pose prediction in the RHD dataset. We also compare the PCK curve obtained by our method with the existing works. Figure \ref{fig:data_corruption_epe_and_baseline_comparison} (third from the left) shows a comparison with the existing works on the RHD dataset. The encoder-decoder models trained using gPoE show comparable performance on unimodal prediction when compared to the baseline Yang et al. \cite{Yang_2019_ICCV}.

\begin{table*}[]
\begin{tabular}{ccccccc}
\toprule
\multirow{2}{*}{Modality Fusing Mechanism} & \multicolumn{2}{c}{No Pixel Corruption}                                  & \multicolumn{2}{c}{25\% Pixel Corruption}                                & \multicolumn{2}{c}{50\% Pixel Corruption}                                \\
\cmidrule{2-7}
                                          & \begin{tabular}[c]{@{}c@{}}mean EPE\\ (mm)\end{tabular} & AUC            & \begin{tabular}[c]{@{}c@{}}mean EPE\\ (mm)\end{tabular} & AUC            & \begin{tabular}[c]{@{}c@{}}mean EPE\\ (mm)\end{tabular} & AUC            \\
\midrule
Mixture-of-Experts (MoE)                   & 12.48                                                   & 0.950          & 14.12                                                   & 0.927          & 15.01                                                   & 0.914          \\
Product-of-Experts (PoE)                   & 12.47                                                   & 0.951          & 13.50                                                   & 0.938          & 15.14                                                   & 0.913          \\
Generalized-Product-of-Experts (gPoE)      & \textbf{12.18}                                          & \textbf{0.955} & \textbf{13.14}                                          & \textbf{0.942} & \textbf{14.73}                                          & \textbf{0.918}\\
\bottomrule
\end{tabular}
  \caption{Comparing gPoE with other modalitity fusing mechanisms for 3d Handpose Estimation from RGB images in RHD \cite{zb2017hand} dataset.}
  \label{tab:modality_fusing_comparison_RHD}
\end{table*}

\begin{table*}[!t]
\begin{tabular}{ccccccc}
\toprule
\multirow{2}{*}{Modality Fusing Mechanism} & \multicolumn{2}{c}{No Pixel Corruption} & \multicolumn{2}{c}{25\% Pixel Corruption} & \multicolumn{2}{c}{50\% Pixel Corruption} \\
\cmidrule{2-7}
                                          & IOU                & F1                 & IOU                 & F1                  & IOU                 & F1                  \\
\midrule
Mixture-of-Experts (MoE)                   & 0.4477             & 0.6143             & 0.4598              & 0.6260              & 0.4545              & 0.6208              \\
Product-of-Experts (PoE)                   & 0.6272             & 0.7692             & 0.6198              & 0.7636              & 0.6221              & 0.7653              \\
Generalized-Product-of-Experts (gPoE)      & \textbf{0.7138}    & \textbf{0.8323}    & \textbf{0.7030}     & \textbf{0.8249}     & \textbf{0.7078}     & \textbf{0.8281}    \\
\bottomrule
\end{tabular}
  \caption{Comparing gPoE with other modalitity fusing mechanisms for predicting segmentation masks from RGB images in Surgical Video-Sim2Real Dataset~\cite{Rivoir_2021_ICCV} dataset.}
  \label{tab:modality_fusing_comparison_surgical}
\end{table*}

\subsection{Modality Fusing Mechanism}

In order to analyze if the proposed modality fusing mechanism plays a significant role in learning unimodal encoders robust to the noise present in the collected dataset, we compare the proposed training mechanism with other existing fusing mechanisms. Since our method builds upon cross-modal VAE architecture, we consider two widely favored methods for fusing information in the latent space, Product-of-Experts, and Mixture-of-Experts. For a fair comparison between different fusing mechanisms, we only change the fusing mechanism module and keep the modality encoders and decoders same in all the settings.

\noindent \textbf{Product-of-Experts (PoE)} proposes a fusing mechanism where the product of the present unimodal posteriors helps formulate the joint posterior. The detailed formulation for PoE can be found in the equations \ref{eq:PoE_p}, \ref{eq:PoE_mu} and \ref{eq:PoE_sigma}.  For an unbiased comparison, we use the learning objective similar to our architecture (equation \ref{eq:learning_objective}), the only difference being the method of computing joint posterior.

\noindent \textbf{Mixture-of-Experts (MoE)} as proposed in mixture-of-experts multimodal variational autoencoder (MMVAE) \cite{shi2019variational}
, factorizes joint posterior by weighted averaging of individual posteriors. Rather than learning individual weights for the modalities, MMVAE suggests giving equal weightage to each modality present to avoid a dominant-modality issue as in PoE. More details can be found in MMVAE \citet{shi2019variational}. Keeping the training strategy same, the learning objective (ELBO) for MoE as proposed in MMVAE is formulated as follows:

\begin{equation}
\begin{aligned}
    \mathrm{ELBO} 
    &=
    \frac{1}{ \mathbf{M}} \sum_{i}^{ \mathbf{M}} \mathbb{E}_{\mathbf{z}_{\mathbf{m}_i} \sim q_{\phi_{\mathbf{m}_i}}\left(\mathbf{z} \mid \mathbf{m}_{i}\right)}\left[\log \frac{p_{\Theta}\left(\mathbf{z}_{\mathbf{m}_i}, \mathbf{m}_{1:  \mathbf{M}}\right)}{q_{\Phi}\left(\mathbf{z}_{\mathbf{m}_i} \mid \mathbf{m}_{1:  \mathbf{M}}\right)}\right] \\
    &=
    \frac{1}{ \mathbf{M}} \sum_{i}^{ \mathbf{M}}
\bigg(
\mathbb{E}_{\mathbf{z}_{m_i} \sim q_{\phi_{\mathbf{m}_i}}\left(\mathbf{z} \mid \mathbf{m}_{i}\right)}\left[\log p_{\Theta}\left(\mathbf{m}_{1:  \mathbf{M}} \mid \mathbf{z}_{\mathbf{m}_i}\right)\right]
\\ &\phantom{xxxxxxxxxx}
-
\operatorname{KL}\left[q_{\Phi}\left(\mathbf{z}_{\mathbf{m}_i} \mid \mathbf{m}_{1:  \mathbf{M}}\right) \| p(\mathbf{z})\right]
\bigg) 
\end{aligned}
\end{equation}

Moreover, to capture the effect of noise on various modality fusing mechanisms, we compare them on different degrees of pixel corruption. We choose no corruption as our baseline and compare them against two additional settings, with 25\%  pixel corruption and 50\% pixel corruption. Table \ref{tab:modality_fusing_comparison_RHD} shows the comparison results on the RHD dataset. The lower mean EPE with a high AUC score highlights the performance boost obtained using the proposed generalized-Product-of-Experts as a modality fusing mechanism. We perform the same set of inferences on 
the segmentation mask prediction task in the Surgical Video-Sim2Real dataset. Table \ref{tab:modality_fusing_comparison_surgical} shows the results of the fusing mechanisms on varying degrees of noise. The proposed modality fusing mechanism shows higher IOU and F1 scores in all the settings when compared to PoE and MoE in multiple degrees of noise present in the test set. Both the tables clearly illustrate the significance of the proposed fusing mechanism in learning robust unimodal encoders and decoders. 

A noteworthy advantage of the proposed fusing mechanism is its ability to train unimodal predictors in a noisy dataset with extra computation cost only during training. During inference, the same architecture results in parameters more robust against noisy samples without any computational overhead.

\subsection{Qualitative Results}
\label{sec:qualitative_results}

For assessing the quality of predictions made by gPoE, we compare it with an architecture using standard Product of Experts for mixing the modalities. 
We train the architectures in the same noisy setting using multiple modalities and evaluate them using only the primary input modality (RGB images). Further, the learned latent spaces are tested using the increasing amount of noise in the RGB modality. Using the learned encoders for both the methods, $\mathbf{z}_{rgb}$ is generated and used as input to the learned hand pose decoder. 
Appendix Figure \ref{app:fig_handpose_inference} shows the predicted hand poses using the learned latent space $\mathbf{z}_{rgb}$ in both the architectures. The predicted 3D keypoints are similar for both the architectures when pixel corruption is small. However, as the pixel corruption increases, gPoE outperforms PoE depicting the importance of dynamic scaling in aligning the latent spaces. 
For Surgical Video-Sim2Real dataset $\mathbf{z}_{rgb}$ is used to predict the segmentation using the learned segmentation decoder. Appendix Figure \ref{app:fig_surgical_prediction} shows the comparison between the architectures trained with PoE and gPoE. The predicted segmentations of gPoE are less noisy when compared  PoE, conveying the robustness for training in a noisy setting.

%% file: Conclusion.tex
\section{Conclusion}

This paper addresses the problem of mixing modalities in a real-world setting where training data captured from one or more modalities are noisy. We propose a novel method for multimodal representation learning in a noisy environment via the generalized product of experts. We test our architecture on two publicly available datasets, Rendered Hand Pose Dataset (RHD) and Surgical Video-Sim2Real, for unimodal prediction tasks in different domains. We show our method's effectiveness in a noisy setting by qualitative and quantitative analysis on various noise levels. 
We observe that the architecture trained by our method leads to unimodal encoders that are more robust toward noisy data samples. Though the proposed method can be generalized to fuse information from N different modalities, we have only tested the architecture considering two input modalities. In future work, we plan to test our architecture on prediction tasks where more than two modalities are available.

